\documentclass{article}
\usepackage{float}
\usepackage{multirow}
\usepackage{array}
\usepackage{amssymb,amsmath}

\usepackage[nonatbib,preprint]{neurips_2020}

\usepackage[utf8]{inputenc} 
\usepackage[T1]{fontenc}    
\usepackage{hyperref}       
\usepackage{url}            
\usepackage{booktabs}       
\usepackage{amsfonts}       
\usepackage{nicefrac}       
\usepackage{microtype}      

\usepackage{graphicx} 
\usepackage{caption,subcaption}
\usepackage{overpic}
\usepackage{amsmath}
\usepackage{physics}
\usepackage{color}
\usepackage{wrapfig}
\usepackage[numbers,sort&compress]{natbib}
\usepackage{cleveref}
\usepackage[final]{pdfpages}
\usepackage{xcolor,pifont}
\usepackage[OT2,T1]{fontenc}
\DeclareSymbolFont{cyrletters}{OT2}{wncyr}{m}{n}
\DeclareMathSymbol{\Sha}{\mathalpha}{cyrletters}{"58}

\usepackage{xcolor}

\newcommand{\cmark}{\ding{51}}%
\newcommand{\xmark}{{\textcolor{red}{\ding{56}}}}

\title{Escaping the Gradient Vanishing: Periodic Alternatives of Softmax in Attention Mechanism}

\author{%
	Shulun Wang$^1$\\
	\texttt{slwang9353@163.com}
	\And
	Bin Liu$^2$\\
	\texttt{xhyous@foxmail.com}
	\And Feng Liu$^1$\\
	\texttt{fliu@bjtu.edu.cn}
}

\begin{document}

\maketitle

\vbox{%
	\vskip -0.26in
	\hskip -0.15in
	\hsize\textwidth
	\linewidth\hsize
	\centering
	\normalsize
	$^1$Department of Computer Science, Beijing Jiaotong University, Beijing, China\\
	$^2$Key Laboratory of Deep Oil and Gas, China University of Petroleum (East China), Qingdao, China\\
	\tt\href{https://github.com/slwang9353/Period-alternatives-of-Softmax}{slwang.github.io/Period alternatives/}
	\vskip 0.3in
}

\begin{abstract}
Softmax is widely used in neural networks for multiclass classification, gate structure and attention mechanisms. The statistical assumption that the input is normal distributed supports the gradient stability of Softmax. However, when used in attention mechanisms such as transformers, since the correlation scores between embeddings are often not normally distributed, the gradient vanishing problem appears, and we prove this point through experimental confirmation. In this work, we suggest that replacing the exponential function by periodic functions, and we delve into some potential periodic alternatives of Softmax from the view of value and gradient. Through experiments on a simply designed demo referenced to LeViT, our method is proved to be able to alleviate the gradient problem and yield substantial improvements compared to Softmax and its variants. Further, we analyze the impact of pre-normalization for Softmax and our methods through mathematics and experiments. Lastly, we increase the depth of the demo and prove the applicability of our method in deep structures.
\end{abstract}
\begin{figure}[H]
\centering
\includegraphics[width=0.9\textwidth,height=5.5cm]{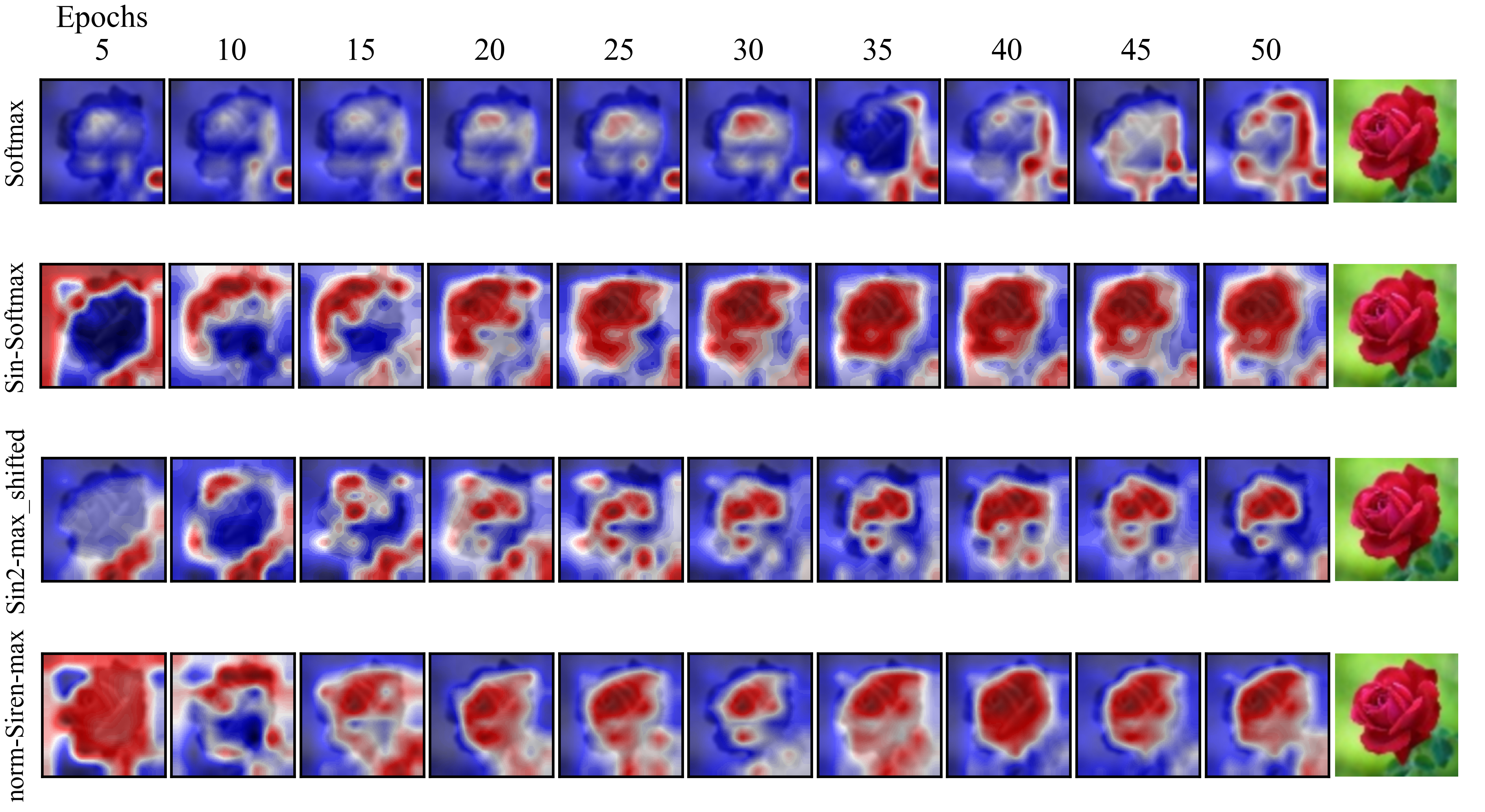}
\caption{Visualization of the attention extracted from transformer blocks. Using Softmax, attention on the boundary is easier to be formed than that on the object, since the score of the boundary is moderate and is less likely to fall into the saturation area, especially in the early stages of training. With our unsaturated periodic alternatives, the attention on the object can be formed smoothly too. Compared to Softmax, our method only needs a short training to form satisfactory attention on the object.}\label{fig1}
\end{figure}

\section{Introduction}
\label{sec:intro}
\begin{figure}[H]
\centering
\includegraphics[width=\textwidth]{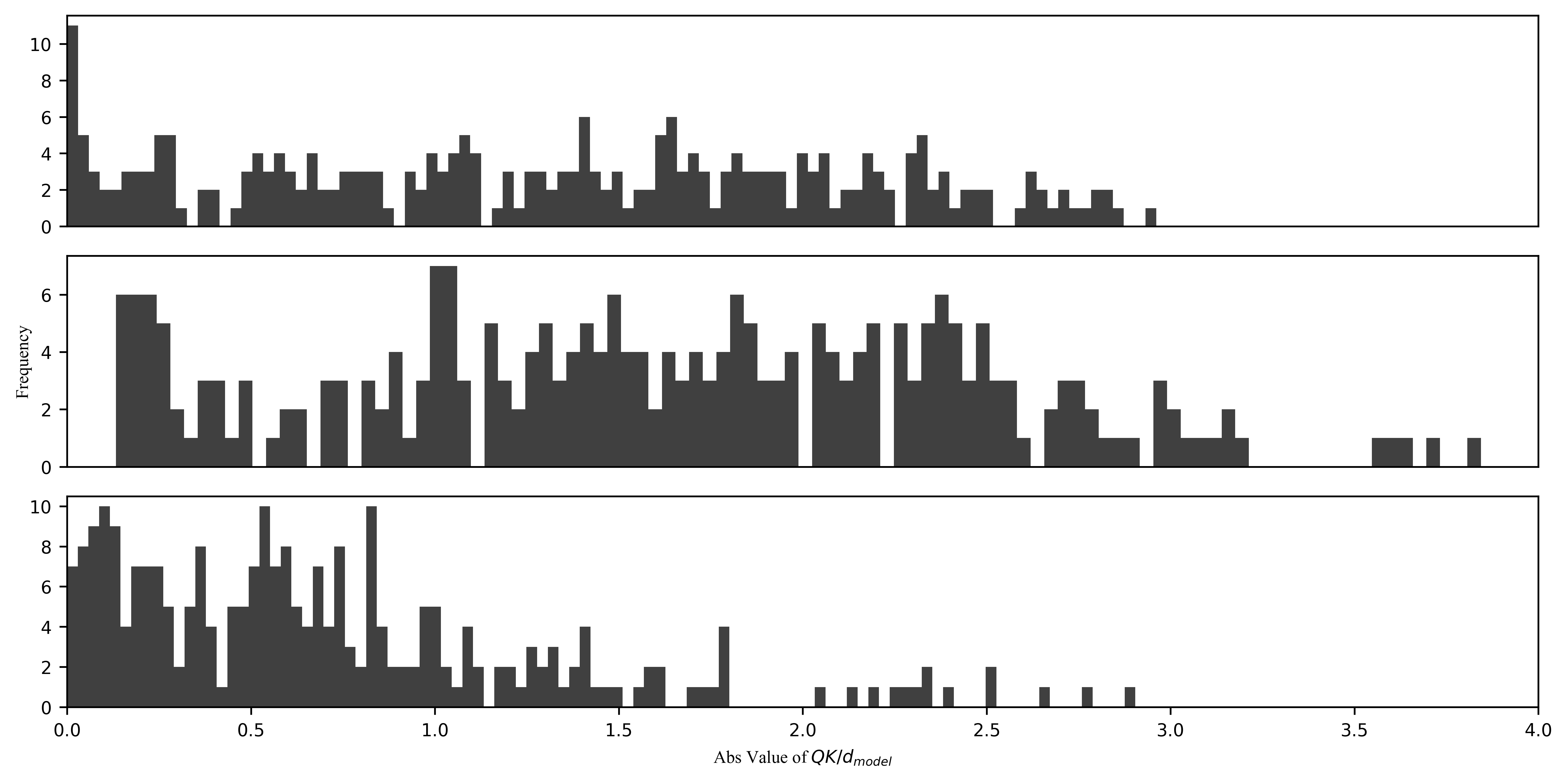}
\caption{Distribution of ${Q} \cdot {K} / d_{model}$ extracted from the transformer blocks. As we speculate, the value of ${Q} \cdot {K}/ d_{model}$ is indeed not strictly normally distributed, and it might fall into the saturation area of Softmax.}\label{fig2}
\end{figure}

{\bf Gradient Vanishing in Attention block:} Since transformer is a pixel-wise process, the distribution of the input is more important than that for CNN. For CNN, the deviation caused by the value exceeding the expected range will be diluted in patches. The statistical assumption that the input is normal distributed supports the gradient stability of Softmax. However, there are always some values that exceed the expected value range, causing the gradient vanishing. This situation becomes worse in the attention mechanisms. The input of Softmax represents to the relationships of embedding (patches), and the distribution of the input should be variant for different images. That means in the attention mechanisms, there are always a part of the input stuck in the saturation area as shown in Figure \ref{fig2}, leading to the gradient vanishing and a long training.

{\bf Motivations:} We are interested in the observation from the transformer-based models that the formation of attention corresponding to the objects often seems to lag behind that corresponding to the boundary, as shown in the first row of Figure \ref{fig1}. Attention can be formed on the boundary in the early stages of training, but slowly appear on objects only in the mid-late stages. However, object should be a more preferred position to put attention on, since there are no inductive bias such as translation equivariance and locality \cite{1}, and objects should be more important than boundary for transformer. By investigating the value of $Q\cdot K$, we find that the value corresponding to the object is larger than boundary, and is more likely to fall into the saturation area of Softmax. Therefore, we speculate that, object is indeed more important, but it is difficult to form attention on object, since the value of $Q\cdot K$ is too large and falls into the saturation area. In contrast, the $Q\cdot K$ value of boundary is moderate, so attention can be formed smoothly. We believe that this situation is one of the reasons why transformer needs a long training.

In this work, we suggest that replacing the exponential function by periodic functions, and we delve into some potential periodic alternatives of Softmax from the view of value and gradient. Through experiments on a simply designed demo referenced to LeViT, our method is proved to be able to alleviate the gradient problem and perform better compared to Softmax and its variants.

To summarize, the main contributions of this paper are:
\begin{itemize}
\item Explore the gradient performance of Softmax in the transformer block, and prove that the input of Softmax is not normal distributed and the gradient vanishing does happens;
\item Introducing a series of periodic alternatives for Softmax in attention mechanisms, which compress the input into the unsaturation area by periodic functions to escape the gradient vanishing;
\item Explore the impact of pre-normalization for Softmax and our methods and make an observation that pre-normalization is just a conditional solution but not always a good choice;
\end{itemize}
\begin{figure}[H]
\centering
\includegraphics[width=0.3\textwidth]{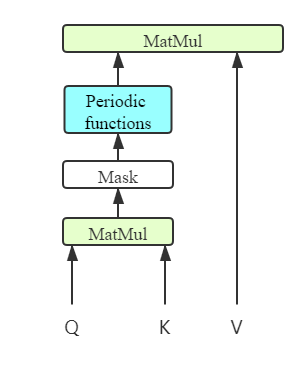}
\caption{The original designed attention block are proposed in \cite{2}. Our work is trying to replace the Softmax with a series of periodic alternatives.}\label{fig3}
\end{figure}

\section{Related works}\label{sec:related}
There are few studies on alternatives of Softmax, since Softmax is mostly used to output classification results, and we can avoid the gradient vanishing by using a joint gradient analytical solution of Softmax and Cross-Entropy. But in the attention mechanism, Softmax is used alone, so the gradient vanishing problem appears. There are other works devoted to enhancing the input feature of Softmax by normalization \cite{3,4,5,6,7}. However, they are all focused on the representation of features, but not addressing the gradient vanishing problem.

{\bf Taylor softmax:} Vincent et al. \cite{8} used second order Taylor series approximation for $e^{x}$ as $1+x+0.5 \cdot x^{2}$ and derive the Taylor softmax as follows:
$$
S_{j}=\frac{1+x_{j}+0.5 \cdot x_{j}^{2}}{\sum_{i}^{d} 1+x_{i}+0.5 \cdot x_{i}^{2}}, x_{j} \subseteq\left(x_{1}, x_{2} \ldots, x_{j}, \ldots x_{d}\right)
$$
where $d$ is the dimension of the input of Taylor softmax. Since the quadratic function changes smoothly, Taylor softmax can generate more soft classification results to alleviate overfitting. However, when used without Cross-Entropy, Taylor softmax will cause gradient vanishing too, because of the saturation area.

{\bf Soft-margin softmax:} Liang et al. \cite{9} introduced a distance margin into Softmax to strengthen the inter-class compactness and inter-class separability between learned features. The Soft-margin (SM) softmax can be described as follows:
$$
S_{j}=\frac{e^{x_{j}-1}}{e^{x_{j}-1}+\sum_{i \neq j}^{d} e_{i}}, x_{j} \subseteq\left(x_{1}, x_{2} \ldots, x_{j}, \ldots x_{d}\right)
$$
where $m$ is manually set and when $m$ is set to zero, the SM-Softmax becomes identical to the original Softmax. SM-Softmax can be considered as a shifted version of Softmax. Similar to Taylor softmax, SM-Softmax and its variant, Ensemble Soft-Margin Softmax \cite{10}, are proposed to encourage the discriminability of features, and the gradient vanishing problem is still not addressed.

{\bf SM-Taylor softmax:} Kunal et al. \cite{11} explored higher order Taylor series approximation of $e^{x}$ to come up with an $n^{ {th }}$ order Taylor softmax where:
$$
f^{n}(x)=\sum_{i=0}^{n} \frac{x^{i}}{i !}
$$
They proved that $f^{n}(x)$ is always positive definite if n is even. Additionally, they combined the strengths of Taylor softmax and Soft-margin softmax, and proposed the SM-Taylor softmax as follows:
$$
S_{j}=\frac{f^{n}\left(x_{j}\right)}{\sum_{i}^{d} f^{n}\left(x_{i}\right)}, x_{j} \subseteq\left(x_{1}, x_{2} \ldots, x_{j}, \ldots x_{d}\right)
$$
However, it is still a method to enhance features, but not a solution to the gradient problem.

\section{Formulation}
For convenience, we denote input, inter-value and scores by $x_{j}$, $f(x)$ and $S_{j}$. In this section, we try to build some periodic functions as the alternatives of Softmax. In addition, there are five aspects to determine whether a function is a favorable alternative: (1) value stability; (2) gradient stability; (3) saturation area; (4) zero-region gradient; (5) information submerged. Furthermore, when judging the gradient-related properties of the functions, we only consider aspects related to $f(x)$ instead of the other elements contained by the input. Since the scores are mapped by $f(x_j)/\sum_{i}^{d} f(x_{j})$, the correlation between the gradient and the other elements is unavoidable for the periodic functions. The plots of $\operatorname{extre}(\partial S_{j} / \partial x_{j})-\sum_{i}^{d} f(x_{j})$ and more discussion on how the other elements in the input influence the gradient of $x_{j}$ are provided in \ref{a1}.

\begin{figure}[H]
\centering
\includegraphics[width=\textwidth]{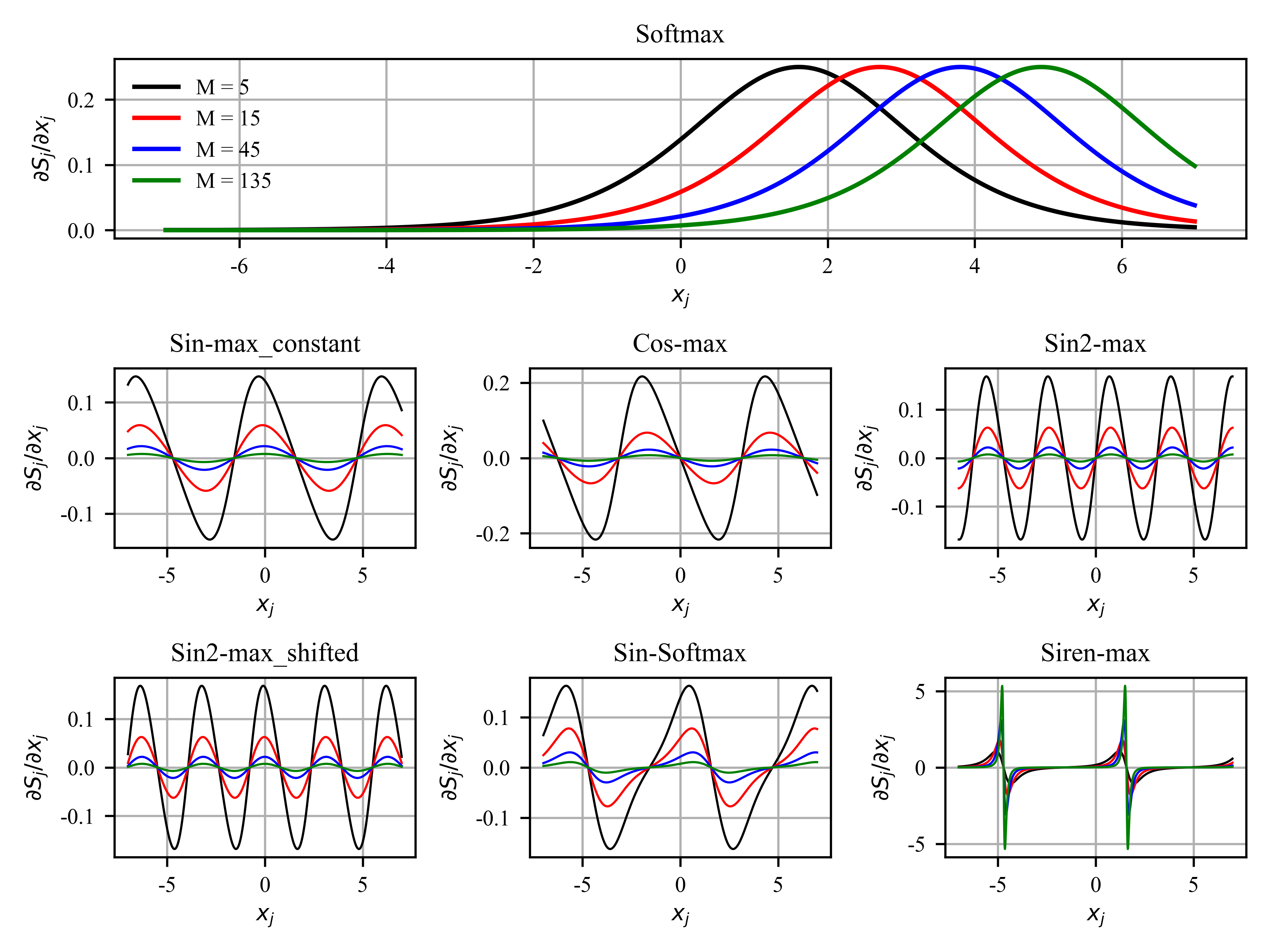}
\caption{Gradient of Softmax and periodic alternatives proposed. $M=\sum_{i \neq j}^{d} f(x)$, and $d$ represents to the dimension of input. Cos-max and Sin2-max have a bad zero-region performance, and there is a saturation area and jump points in Siren-max. Except for them, other periodic alternatives are well performed in zero-region and they are all unsaturated and stable.}\label{fig4}
\end{figure}

According to the research in the Taylor Softmax function proposed by Vincent et al. \cite{8}, and the higher order Taylor Softmax proposed by Kunal et al. \cite{11}, it is reasonable to map the input with a monotonic function, since the input can adapt to a suitable value range as parameters update. Therefore, we suggest to use a periodic function $f(x)$ to compress the input, so as to avoid approximating the small input to a fixed value (to keep them positive), and also avoid the output being too large to have an appropriate gradient.

\subsection{Softmax}
What Softmax does is mapping the input $x$ to an inter-value $f(x)$, $f(x)=e^{x}$, and mapping the inter-value to scores $S_{j}$. The exponential function keeps the negative input positive, but also makes the positive input extremely high. For a large input $x_{j}$, $e^{x_{j}}$ is too large and dominates $S_{j}$, which means $S_{j} \approx 1$, $\partial S_{j} / \partial x_{j}=0$; and for a small input $x_{j}$, $\partial e^{x_{j}} / \partial x_{j}=0$, and since $\partial S_{j} / \partial x_{j}=(\partial S_{j} / \partial e^{x_{j}}) \cdot(\partial e^{x_{j}} / \partial x_{j})$, $\partial S_{j} / \partial x_{j}=0$. Therefore, the major cause of gradient vanishing in Softmax is that we need to compress the value $f(x)$ with unknown upper and lower into $(0,+\infty)$. To do this, there has to be a saturation area for large and small value. 

Before the discussion of the alternatives of Softmax, it is necessary to clarify the advantages of Softmax. First of all, the output of Softmax is positive definite. And due to exponential function, the difference between inputs will be magnified, that means Softmax can show the difference between inputs well, which is a good characteristic for attention mechanisms. Besides, according to the definition:
$$
\begin{gathered}
S_{\text {softmax }_{j}}=e^{x_{j}} / \sum_{i}^{d} e^{x_{i}}, x_{j} \subseteq\left(x_{1}, x_{2} \ldots, x_{j}, \ldots x_{d}\right) \\
\partial S_{j} / \partial x_{j}=M \cdot e^{x_{j}} /\left(M+e^{x_{j}}\right)^{2}, M=\sum_{i \neq j}^{d} e^{x_{i}} \\
\frac{\partial^{2} S_{j}}{\partial x_{j}^{2}}=M \cdot e^{x_{j}} \cdot\left(e^{x_{j}}+M\right)^{-2}-2 \cdot M \cdot e^{2 x_{j}} \cdot\left(e^{x_{j}}+M\right)^{-3}
\end{gathered}
$$

Let $\partial^{2} S_{j} / \partial x_{j}^{2}=0$, we have:
$$
\begin{gathered}
M \cdot e^{x_{j} }\cdot\left(e^{x_{j}}+M\right)^{-2}-2 \cdot M \cdot e^{2 x_{j}\cdot\left(e^{x_{j}}+M\right)^{-3}=0} \\
e^{x_{j}}=\mathrm{M} \\
\qquad \text { Extre }\left(\frac{\partial S_{j}}{\partial x_{j}}\right)=M \cdot M /(M+M)^{2}=1 / 4
\end{gathered}
$$
which means the max gradient of Softmax is 0.25, so it will not cause gradient explosions. Additionally, in spite of the saturation problem, no matter how large the other elements of input are, a sufficiently large value can always get an appropriate gradient.

\subsection{Sin-max-constant / Cos-max:}
When it comes to periodic functions, there is a good reason to use sine function since it is widely used and derivable. Therefore, we set the $f(x)=1+\sin (x)$ to keep the function positive definite following the suggestion in \cite{12}, and Sin-max-constant is defined as follows:
$$S_{\sin\max_{j}}=\frac{1+\sin \left(x_{j}\right)}{d+M+\sin \left(x_{j}\right)}, M=\sum_{i \neq j}^{d} \sin \left(x_{i}\right)$$
where $d$ represents to the dimension of input. Sin-max compress $x$ into $(0,2)$ and for $\sin(x)$, there is no saturation area.

However, let $x \sim N(0,1)$, then $E(M) \approx-\sin \left(x_{j}\right)$, we have:
$$
E(S_{\sin\max_{j}})=\frac{1+\sin (x_{j})}{d}=\frac{1}{d}+\frac{\sin (x_{j})}{d}, d \gg \sin (x_{j}) 
$$
$$
E(S_{\sin\max_{j}}) \approx \frac{1}{d}
$$
which means as the dimension of $x$ increases, the influence of $x$ will be weakened by the constant term, causing $x$ to be overwhelmed and cannot be mapped to $S$ correctly. Besides, consider from the view of gradient:
$$
\frac{\partial S_{j}}{\partial x_{j}}=\frac{(M+d-1) \cdot \cos \left(x_{j}\right)}{\left(M+d+\sin \left(x_{j}\right)\right)^{2}} 
$$
$$
E\left(\frac{\partial S_{j}}{\partial x_{j}}\right)=\frac{\left(-\sin \left(x_{j}\right)+d-1\right) \cdot \cos \left(x_{j}\right)}{d^{2}}
$$
Similar to the $E(S\sin\max_{j})$, as the dimension of $x$ increases, the gradient of Sin-max will drop to zero, and gradient vanish will occur in the entire value range. The main reason for these defects is the constant term 1 in $f(x)$.

We try to remove the constant term in $f(x)$, and the expression becomes to:
$$
S_{\sin\max_{j}}=\frac{\sin \left(x_{j}\right)}{M+\sin \left(x_{j}\right)},  M=\sum_{i \neq j}^{d} \sin \left(x_{i}\right)
$$

Now let $x \sim N(0,1)$, then $E\left(M+\sin \left(x_{i}\right)\right)=0$, we have:
$$E(S_{\sin\max_{j}})=\frac{\sin(x_{j})}{0} \rightarrow \pm \infty$$

which means the value of $S_{{\sin\max}_{j}}$ is unstable, causing the network to have a great risk of
breaking down. And for gradient, since $E(M+\sin (x_{j}))=0$ and $E(M)=-\sin(x_{j})$, we have:
$$
\frac{\partial S_{j}}{\partial x_{j}}=\frac{M \cdot \cos \left(x_{j}\right)}{\left(M+\sin \left(x_{j}\right)\right)^{2}}
$$
$$
E\left(\frac{\partial S_{j}}{\partial x_{j}}\right)=\frac{-0.5 \cdot \sin \left(2 \cdot x_{j}\right)}{0} \rightarrow \pm \infty
$$
which is also unstable because $f(x)$ is not positive definite.

The reason why $f(x)=1+\cos (x)$ is not good is similar to Sin-max-constant where the difference within $x$ will be submerged due to the constant term as the dimension growing. Let $f(x)=\cos (x)$, and $x \sim N(0,1)$, we can define Cos-max as:

$$
\begin{gathered}
S_{\cos\max_{j}}=\frac{\cos \left(x_{j}\right)}{M+\cos \left(x_{j}\right)}, M=\sum_{i \neq j}^{d} \cos \left(x_{i}\right) \\
\frac{\partial S_{j}}{\partial x_{j}}=\frac{-M \cdot \sin \left(x_{j}\right)}{\left(M+\cos \left(x_{j}\right)\right)^{2}}
\end{gathered}
$$

Assume that $\mathrm{x}$ strictly belongs to $[-1,1]$, then Cos-max can be considered as the Sin-max shifted to a positive definite range. However, a gradient vanishing problem will appear when $x$ clusters in the zero-region. Besides, from the view of gradient stability, we have:

$$
\frac{\partial^{2} S_{j}}{\partial x_{j}^{2}}=-M \cdot \cos \left(x_{j}\right) \cdot\left(M+\cos \left(x_{j}\right)\right)^{-2}+2 \cdot M \cdot \sin \left(x_{j}\right) \cdot\left(M+\cos \left(x_{j}\right)\right)^{-3}
$$

Let $\partial^{2} S_{j} / \partial x_{j}^{2}=0$, we have:

$$
\cos \left(x_{j}\right)-2 \cdot \sin \left(x_{j}\right) \cdot\left(M+\cos \left(x_{j}\right)\right)^{-1}=0 
$$
$$
-4+\left(M^{2}+1\right) \cdot \cos ^{2}\left(x_{j}\right)+2 \cdot M \cdot \cos^{3}\left(x_{j}\right)+\cos ^{4}\left(x_{j}\right)=0
$$

According to the solving provided in \ref{a5}, we can have:
$$
\operatorname{Extre}\left(\frac{\partial S_{j}}{\partial x_{j}}\right)_{M} \in(-\infty,+\infty)
$$

As $\mathrm{M}$ increases, Extreme $\left(\partial S_{j} / \partial x_{j}\right)$ approaches $\pm \infty$, which means Cos-max is gradient unstable, causing the network to have a great risk of breaking down like Sin-max.

\subsection{Sin2-max-shifted}
To ensure that $f(x)$ is positive definite, and no extra constant term is introduced, $\sin ^{2} x$ is a reasonable choice. So we can define Sin2-max as:
$$
S_{\sin2\max_j}=\frac{\sin ^{2}\left(x_{j}\right)}{M+\sin ^{2}\left(x_{j}\right)}, M=\sum_{i \neq j}^{d} \sin ^{2}\left(x_{i}\right) $$
$$
\frac{\partial S_{j}}{\partial x_{j}}=\frac{M \cdot \sin \left(2 \cdot x_{j}\right)}{\left(M+\sin ^{2}\left(x_{j}\right)\right)^{2}}
$$
Note that although $\sin ^{2}(x)=1 / 2-\cos (2 \cdot x) / 2, f_{\sin 2 \max }(x)=1 / 2-f_{\cos\max}(2 \cdot x) / 2$, Sin2-max is not just a scaled double-frequency version of Cos-max owing to the constant terms. Therefore, the numerical and gradient characteristics of Sin2-max and Cos-max are different.

As shown in Figure \ref{fig4}, the possible problem of Sin2-max is that, assuming $x \sim N(0,1)$, most of $x$ clusters in the region close to 0 so most of the gradients close to 0, which makes the parameters difficult to update. To solve this `conditional' problem, we can shift $\sin ^{2}(x)$ to the non-zero region by adding a phase $\varphi$ to $x$.

Let $\partial^{2} S_{j} / \partial x_{j}^{2}=0 .$ We have:
$$
\cos \left(2 \cdot x_{j}\right)=-\frac{1}{2} \cdot\left(2 \cdot M+1 \pm \sqrt{8+(2 \cdot M+1)^{2}}\right) 
$$
$$
x_{j} \text { for } \max \left(\frac{\partial S_{j}}{\partial x_{j}}\right)=\frac{1}{2} \operatorname{arcos}\left(-\frac{1}{2} \cdot\left(2 \cdot M+1 \pm \sqrt{8+(2 \cdot M+1)^{2}}\right)\right)
$$

Unfortunately, $x_{j}$ for $\max \left(\partial S_{j} / \partial x_{j}\right)$ will change with $M$, so we have to find an approximate
solution. Besides, with $M$ change, the gradient will oscillate in the range of $(0,+\infty)$, causing gradient explosion or vanishing in the entire value range. Since $\sin \left(2 \cdot x_{j}\right)$ and $\sin ^{2}\left(x_{j}\right)$ has a same cycle of period $\pi$, the period of $\partial S_{j} / \partial x_{j}$ is $\pi$. We set $\varphi=\pi / 4$ to make the gradient stable. So we get Sin2-max-shifted as follows:
$$
S_{\sin2\text{max-shifted}_{j}}=\frac{\sin ^{2}\left(x_{j}+0.25 \cdot \pi\right)}{M+\sin ^{2}\left(x_{i}+0.25 \cdot \pi\right)},  M=\sum_{i \neq j}^{d} \sin ^{2}\left(x_{i}+0.25 \cdot \pi\right)
$$
$$
\frac{\partial S_{j}}{\partial x_{j}}=\frac{M \cdot \sin \left(2 \cdot x_{j}+0.5 \cdot \pi\right)}{\left(M+\sin ^{2}\left(x_{i}+0.25 \cdot \pi\right)\right)^{2}}
$$
\subsection{Sin-Softmax}
From another view, instead of replacing the exponential function, compressing the input into the unsaturation area is also a reasonable choice. To keep the gradient of input in zero region away from 0, we choose sine but not cosine. We define Sin-Softmax as follows:
$$
S_{\text {sin-softmax}_{j}}=\frac{e^{\sin \left(x_{j}\right)}}{M+e^{\sin \left(x_{j}\right)}},  M=\sum_{i \neq j}^{d} e^{\sin \left(x_{i}\right)}
$$
$$
\frac{\partial S_{j}}{\partial x_{j}}=\frac{M \cdot e^{\sin \left(x_{j}\right)} \cdot \cos \left(x_{j}\right)}{\left(M+e^{\sin \left(x_{j}\right)}\right)^{2}}
$$
Sin-Softmax can be considered as a periodic-normalized version of Softmax, which is also similar to the periodic activation function proposed in SirenNet \cite{13}. The best part of Sin-Softmax is that, the input is compressed into the well performed region of Softmax by periodic function, so the value and gradient are both stable, as show in Figure \ref{fig4} and \ref{a1}. Additionally, $S_j$ is positive definite owing to the exponential function, and the gradient in the zero-near region is also well performed. However, the possible defect of Sin-Softmax is that, the largest score can only be $e^{2}$ times the smallest, since for $x_{j} \in R:$
$$
\sin \left(x_{j}\right) \in(-1,+1),  e^{\sin \left(x_{j}\right)} \in\left(\frac{1}{e}, e\right)
$$
which might cause oblivion of the most contributing value covered in a large number of low contributing values, as the dimension of the score maps (or the number of embeddings) increasing.

\subsection{Siren-max}
Inspired by SirenNet \cite{13} where sine is used as the activation function, we define a $f(x)$ with benefic gradient properties by:
$$
f(x)=\frac{\sin (x)}{1-\sin (x)}
$$
Since $f(x) \in[-0.5,+\infty)$, to make $S_{j}$ positive definite, we add $0.5$ to $f(x)$ and define Siren-max as follows:
$$
S_{\text {siren-max }_{j}}=\frac{1+\sin \left(x_{j}\right)}{2-2 * \sin \left(x_{j}\right)} /\left(M+\frac{1+\sin \left(x_{j}\right)}{2-2 * \sin \left(x_{j}\right)}\right), M=\sum_{i \neq j}^{d} \frac{1+\sin \left(x_{i}\right)}{2-2 * \sin \left(x_{i}\right)}
$$
$$
\frac{\partial S_{j}}{\partial x_{j}}=\frac{4 \cdot M \cdot \cos \left(x_{j}\right)}{\left(2 \cdot M-2 \cdot M \cdot \sin \left(x_{j}\right)+1+\sin \left(x_{j}\right)\right)^{2}}
$$
Note that, the upper bound of $f(x)$ is infinity, so adding a constant term to it will not cause the difference between $x$ being submerged like Sin-max-constant. As shown in Figure \ref{fig4}, there is no saturation area in Siren-max and it is well performed in zero-near region. The possible defect is that, the gradient has periodic jump points which might make training unstable.

\begin{table}[H]
\caption{A summary of Softmax and the periodic alternatives. Note that the judgment is based on the assumption that the input is standard normal distributed, and \ding{56}/\ding{51} means that when input exceed the expected value range, which always happens in attention mechanisms, the indicators go bad.}\label{tb1}
\resizebox{\textwidth}{!}{
\begin{tabular}{lp{2cm}<\centering p{2cm}<\centering p{2cm}<\centering p{2cm}<\centering p{2cm}<\centering}
\toprule
                          & \textbf{Value stability?} & \textbf{Gradient stability?} & \textbf{No saturation area?} & \textbf{Zero-region gradient good?} & \textbf{Info Submerged?}\\
                          \midrule
                          \textbf{Softmax}          & \cmark                         & \cmark                            & \textcolor{red}{\ding{56}}/\ding{51}                          & \cmark                                   & \cmark                        \\\textbf{Sin-max-constant} & \xmark                         & \cmark                            & \cmark                            & \cmark                                   & \xmark                        \\\textbf{Cos-max}          & \textcolor{red}{\ding{56}}/\ding{51} & \textcolor{red}{\ding{56}}/\ding{51}                          & \cmark                            & \xmark                                   & \cmark                        \\\textbf{Sin2-max}         & \cmark                         & \cmark                            & \cmark                            & \xmark                                   & \cmark                        \\\textbf{Sin2-max-shifted} & \cmark                         & \cmark                            & \cmark                            & \cmark                                   & \cmark                        \\\textbf{Sin-Softmax}      & \cmark                         & \cmark                            & \cmark                            & \cmark                                   & \cmark                        \\
                          \textbf{Siren-max}        & \cmark                         & \textcolor{red}{\ding{56}}/\ding{51}                          & \cmark                            & \cmark                                   & \cmark                      
                          \\\bottomrule \end{tabular}}
\end{table}

\subsection{Pre-normalization}
Note that the pre-normalization discussed is not parameterized like Batch-norm \cite{14}, Layer-norm \cite{15} or Group-norm \cite{16}. We denote the pre-normalization by normalizing the elements of $Q \cdot K$ in row. Considering the saturation problem, normalizing the input is also a reasonable operation. However, since the distribution of attention score maps is different for variant images, the normalization can hardly compress the maps into a specified value range precisely. Besides, the function $f(x_{j})={normlized}\left(x_{j}\right)$ also might be saturated, and the saturation area shifts with $\mathrm{E}(x_{j})$ and $\operatorname{Var}(x_{j})$. Therefore, the gradient situation of ${normlized}(x_{j})$ is similar to that of Softmax, which may cause gradient vanishing too. As a result, although pre-normalization can roughly gather the values in a specified range, but on the contrary, it may bring new gradient problems. More discussion and the gradient plots of normalization, pre-normalized version of Softmax and periodic alternatives are provide in \ref{a2}. 

\section{Experiments}
\begin{figure}[H]
\centering
\includegraphics[width=0.8\textwidth]{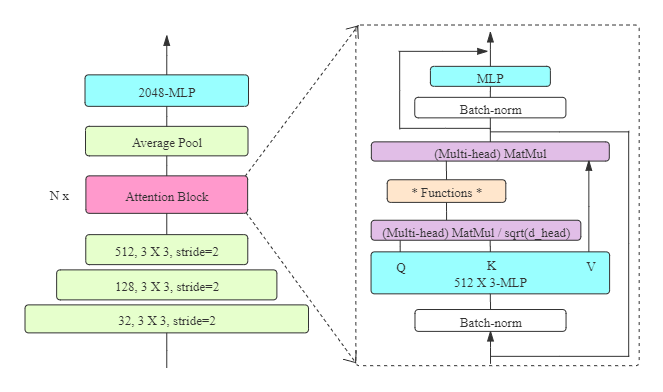}
\caption{The structure of the demo trained in experiments. Except for the number of attention blocks and the number of neurons of the classification head, the other parts remain unchanged in the experiment.}\label{fig5}
\end{figure}

To eliminate the unexpected effects of various tricks, the experiments are operated on a simply designed demo referenced to LeViT \cite{1}, as shown in Figure \ref{fig5}. 

\begin{figure}[H]
\centering
\includegraphics[width=\textwidth]{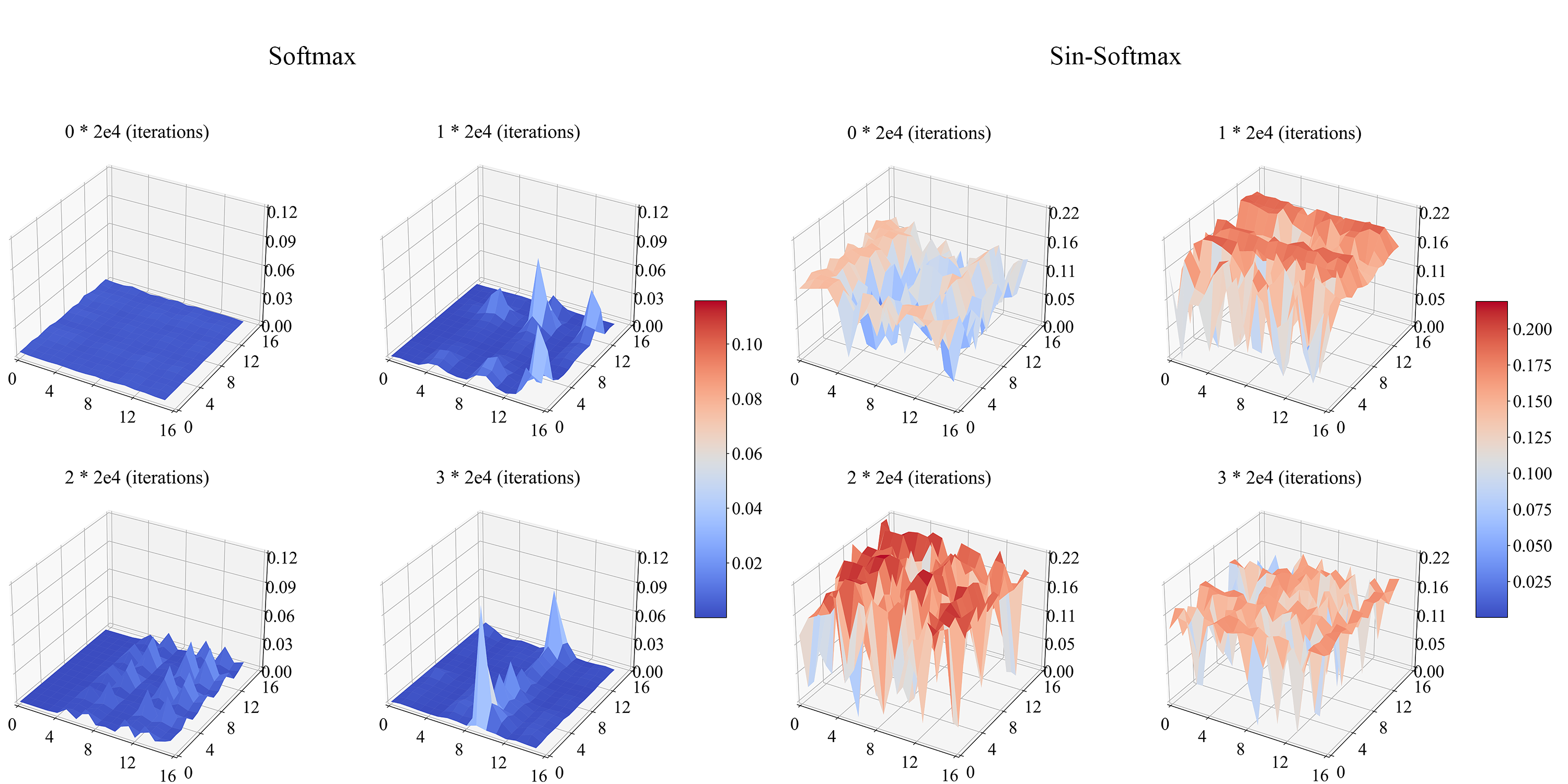}
\caption{3D graph of the gradient extracted from attention blocks under Softmax and Sin-Softmax. The input of Softmax represents the correlation of embedding (patches), and the distribution of correlation should be variant for different images. As a result, the distribution of $K$, $Q$ will gradually deviate from the normal distribution, as will the input of Softmax. Since a part of the input might fall into the saturation area, gradient vanishing occurs. And our methods successfully address this problem by introducing the period functions to generate attention scores. More 3D gradient graphs of our methods are provided in \ref{a3}.}\label{fig6}
\end{figure}

In experiments, there is an observation that most of the gradient of Softmax are very small, and only a small part of updates can be successfully back-propagated, even in the early stages of training, as shown in Figure \ref{fig6}. This phenomenon proves our point that the $Q\cdot K$ value, which are used to generate the attention scores, are related to the input images content and are not strictly normal distributed. Therefore, even if divided by $\sqrt{d}$ following the original designed transformer block, the value still might fall into the saturation area of Softmax, making updates difficult. In our method, there is no saturation area in the functions, so the gradient is satisfactory at each training stage, which promotes the updating of parameters. More 3D graphs of gradients extracted from experiment are provide in \ref{a3}.

\begin{figure}[H]
\centering
\includegraphics[width=0.8\textwidth]{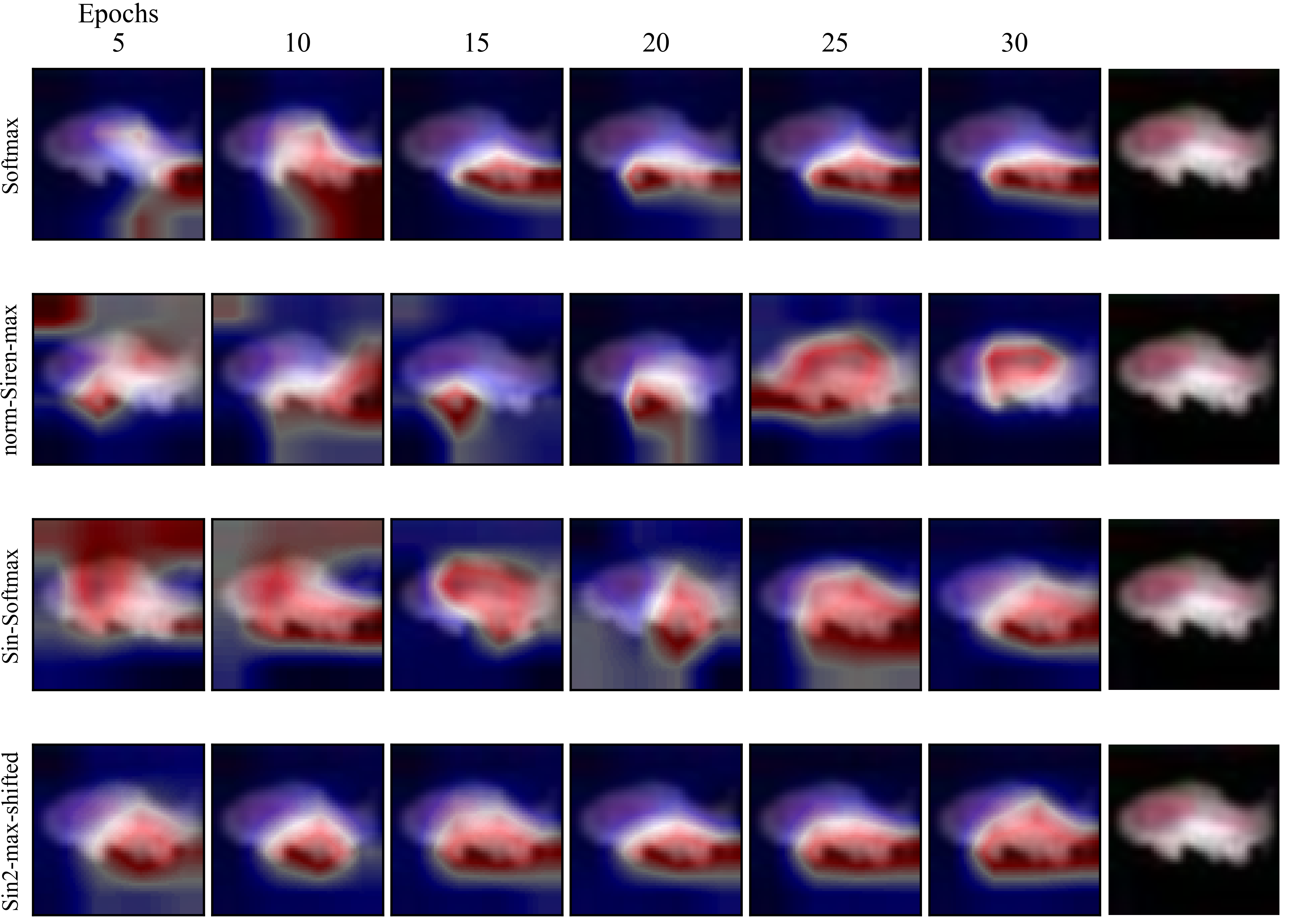}
\caption{Visualization of the attention extracted from transformer blocks. In softmax, attention is more likely to be trapped on the boundary, since the $QK$ value of the boundary is moderate and not easy to fall into the saturation area. By contrast, our method eliminates the saturation area, and attention can be updated smoothly on the objects too.}\label{fig7}
\end{figure}

As shown in Figure \ref{fig1} and Figure \ref{fig7}, due to the gradient vanishing problem, Softmax might cause difficulty in the formation of attention, especially in the early stages of training. We observe that attention is formed more smoothly on the boundary, and on the contrary, the attention corresponding to the objects can only be formed in the later stages of training. A possible reason is that the scores of the boundary between object and content are moderate, so the gradient flows smoothly. While, the scores of the objects are larger and might fall into the saturation area of Softmax, causing the gradient vanishing and the formation of attention being locked. While under the periodic alternatives, the attention is updated unrestrained on the image, which strengthens our arguments.

\begin{table}[H]
\centering
\caption{Comparison with Softmax under same model with increasing depth. We also report pre-normalization version result of each functions. Note that, since the trainings under Sin-max-constant, Cos-max, Sin2-max breaks down in the early stages, we omit their results here. In addition, the depth represents to the number of attention blocks stacked, and the rest structures of demo remain unchanged.}\label{tb2}
\begin{tabular}{ccccccccc}
\toprule
\multirow{2}{*}{\textbf{\begin{tabular}[c]{@{}c@{}}Structure = Demo\\ Dataset = Cifar-100\end{tabular}}} & \multicolumn{2}{c}{\textbf{Depth = 1}} & \multicolumn{2}{c}{\textbf{Depth = 2}} & \multicolumn{2}{c}{\textbf{Depth = 4}} & \multicolumn{2}{c}{\textbf{Depth = 8}} \\
\cline{2-9}
                                                                                                         &                         & norm         &                         & norm         &                     & norm             &                         & norm         \\ \midrule
\textbf{Softmax}                                                                                         & 52.67                   & 53.80        & 58.21                   & 59.57        & 67.01               & 68.75            & 81.12                   & 82.31        \\
\textbf{Sin2-max-shifted}                                                                                & 53.90                   & 53.16        & 60.37                   & 59.84        & 71.48               & 72.38            & 84.81                   & 83.20        \\
\textbf{Sin-Softmax}                                                                                     & \textbf{54.64}          & 54.27        & \textbf{60.40}          & 60.18        & 72.39               & 72.84            & \textbf{85.14}          & 84.63        \\
\textbf{Siren-max}                                                                                       & \textbackslash{}        & 55.19        & \textbackslash{}        & 59.62        & \textbackslash{}    & \textbf{73.25}   & \textbackslash{}        & 84.70        \\\bottomrule
\multicolumn{9}{c}{\textbf{*  \textbackslash ~means training breaks down in the early stages}}                                                                                                                                                                              
\end{tabular}
\end{table}

The gradient performance in the zero-region is crucial for training, and the early breaking down of training under Cos-max and Sin2-max can be ascribed to this. Besides, the stability of gradient is also very important. Since there are jump points within the range of input, the training under Siren-max breaks down too. In addition, since the input is submerged in the constant terms, the training under Sin-max-constant diverges.

Encouragingly, Sin2-max-shifted, Sin-Softmax exceed Softmax in the result just as we speculate, and norm-Siren-max is also surprisingly well performed. The result are shown in Table \ref{tb2}. The major drawbacks of Cos-max and Sin-max-constant are gradient performance in zero-region and information submergence respectively, which cannot be optimized by pre-normalization. As for Siren-max, pre-normalization optimizes the distribution of input and helps Siren-max avoid the gradient jump points, resulting in a satisfactory performance. Softmax can also be improved since pre-normalization helps the input escape from the saturation area to some degree. However, Sin2-max-shifted and Sin-Softmax are not subject to input distribution, so they cannot get benefits from pre-normalization. On the contrary, since pre-normalization will bring unexpected gradient problems, the performance of norm-Sin2-max-shifted and norm-Sin-Softmax decrease slightly. The plots and complete results of the experiment are provided in \ref{a4}.

\section{Conclusion}
Through the visualization of attention and gradient extracted from transformer blocks, we prove that in the attention mechanism, Softmax does lead to the gradient vanishing problem and makes training difficult. To address the problem, we propose a series of periodic alternatives of Softmax, and the experimental results prove that Sin-Softmax, Sin2-max-shifted, and norm-Siren-max are better performed than Softmax in attention mechanism. Additionally, we make an observation that pre-normalization is just a conditional solution but not always a good choice. 

In the periodic alternatives, embedding requiring more attention does not necessarily require a larger $Q\cdot K$ value, which makes the generation of $Q$, $K$ more free, and it is hard to say whether this will lead to unexpected problems. This change might affect the representation of the model, and we will explore how this change happens in further works.

{\small
	\bibliographystyle{unsrtnat}
	\bibliography{main}

\begin{thebibliography}{16}
\providecommand{\natexlab}[1]{#1}
\providecommand{\url}[1]{\texttt{#1}}
\expandafter\ifx\csname urlstyle\endcsname\relax
  \providecommand{\doi}[1]{doi: #1}\else
  \providecommand{\doi}{doi: \begingroup \urlstyle{rm}\Url}\fi

\bibitem[Dosovitskiy et~al.(2021)Dosovitskiy, Beyer, Kolesnikov, Weissenborn,
  Zhai, Unterthiner, Dehghani, Minderer, Heigold, Gelly, et~al.]{1}
Alexey Dosovitskiy, Lucas Beyer, Alexander Kolesnikov, Dirk Weissenborn,
  Xiaohua Zhai, Thomas Unterthiner, Mostafa Dehghani, Matthias Minderer, Georg
  Heigold, Sylvain Gelly, et~al.
\newblock An image is worth 16x16 words: Transformers for image recognition at
  scale.
\newblock In \emph{Proc. ICLR}, 2021.

\bibitem[Vaswani et~al.(2017)Vaswani, Shazeer, Parmar, Uszkoreit, Jones, Gomez,
  Kaiser, and Polosukhin]{2}
Ashish Vaswani, Noam Shazeer, Niki Parmar, Jakob Uszkoreit, Llion Jones,
  Aidan~N Gomez, {\L}ukasz Kaiser, and Illia Polosukhin.
\newblock Attention is all you need.
\newblock \emph{arXiv preprint arXiv: 1706.03762v5}, 2017.

\bibitem[Liu et~al.(2016)Liu, Wen, Yu, and Yang]{3}
Weiyang Liu, Yandong Wen, Zhiding Yu, and Meng Yang.
\newblock Large-margin softmax loss for convolutional neural networks.
\newblock In \emph{ICML}, volume~2, page~7, 2016.

\bibitem[Wang et~al.(2017)Wang, Xiang, Cheng, and Yuille]{4}
Feng Wang, Xiang Xiang, Jian Cheng, and Alan~Loddon Yuille.
\newblock Normface: {$L_2$} hypersphere embedding for face verification.
\newblock \emph{arXiv preprint arXiv: 1704.06369v4}, 2017.

\bibitem[Ranjan et~al.(2017)Ranjan, Castillo, and Chellappa]{5}
Rajeev Ranjan, Carlos~D Castillo, and Rama Chellappa.
\newblock {$L_2$}-constrained softmax loss for discriminative face
  verification.
\newblock \emph{arXiv preprint arXiv:1703.09507v3}, 2017.

\bibitem[Liu et~al.(2018)Liu, Wen, Yu, Li, Raj, and Song]{6}
Weiyang Liu, Yandong Wen, Zhiding Yu, Ming Li, Bhiksha Raj, and Le~Song.
\newblock Sphereface: Deep hypersphere embedding for face recognition.
\newblock \emph{arXiv preprint arXiv: 1704.08063v4}, 2018.

\bibitem[Zheng et~al.(2018)Zheng, Pal, and Savvides]{7}
Yutong Zheng, Dipan~K Pal, and Marios Savvides.
\newblock Ring loss: Convex feature normalization for face recognition.
\newblock \emph{1803.00130v1}, 2018.

\bibitem[Vincent et~al.(2014)Vincent, De~Br{\'e}bisson, and Bouthillier]{8}
Pascal Vincent, Alexandre De~Br{\'e}bisson, and Xavier Bouthillier.
\newblock Efficient exact gradient update for training deep networks with very
  large sparse targets.
\newblock \emph{arXiv preprint arXiv:1412.7091v1}, 2014.

\bibitem[Liang et~al.(2017)Liang, Wang, Lei, Liao, and Li]{9}
Xuezhi Liang, Xiaobo Wang, Zhen Lei, Shengcai Liao, and Stan~Z Li.
\newblock Soft-margin softmax for deep classification.
\newblock In \emph{Proc. ICONIP}, pages 413--421. Springer, 2017.

\bibitem[Wang et~al.(2018)Wang, Zhang, Lei, Liu, Guo, and Li]{10}
Xiaobo Wang, Shifeng Zhang, Zhen Lei, Si~Liu, Xiaojie Guo, and Stan~Z Li.
\newblock Ensemble soft-margin softmax loss for image classification.
\newblock \emph{arXiv preprint arXiv:1805.03922v1}, 2018.

\bibitem[Banerjee et~al.(2020)Banerjee, Gupta, Vyas, Mishra, et~al.]{11}
Kunal Banerjee, Rishi~Raj Gupta, Karthik Vyas, Biswajit Mishra, et~al.
\newblock Exploring alternatives to softmax function.
\newblock \emph{arXiv preprint arXiv:2011.11538v1}, 2020.

\bibitem[de~Br{\'e}bisson and Vincent(2016)]{12}
Alexandre de~Br{\'e}bisson and Pascal Vincent.
\newblock An exploration of softmax alternatives belonging to the spherical
  loss family.
\newblock \emph{arXiv preprint arXiv:1511.05042v3}, 2016.

\bibitem[Sitzmann et~al.(2020)Sitzmann, Martel, Bergman, Lindell, and
  Wetzstein]{13}
Vincent Sitzmann, Julien Martel, Alexander Bergman, David Lindell, and Gordon
  Wetzstein.
\newblock Implicit neural representations with periodic activation functions.
\newblock \emph{arXiv preprint arXiv: 2006.09661v1}, 2020.

\bibitem[Ioffe and Szegedy(2015)]{14}
Sergey Ioffe and Christian Szegedy.
\newblock Batch normalization: Accelerating deep network training by reducing
  internal covariate shift.
\newblock \emph{arXiv preprint arXiv: 1502.03167v3}, 2015.

\bibitem[Ba et~al.(2016)Ba, Kiros, and Hinton]{15}
Jimmy~Lei Ba, Jamie~Ryan Kiros, and Geoffrey~E Hinton.
\newblock Layer normalization.
\newblock \emph{arXiv preprint arXiv:1607.06450v1}, 2016.

\bibitem[Wu and He(2018)]{16}
Yuxin Wu and Kaiming He.
\newblock Group normalization.
\newblock \emph{arXiv preprint arXiv: 1803.08494v3}, 2018.

\end{thebibliography}
}

\newpage
\appendix
\setcounter{section}{1}
\begin{center}
\Large{\bf Escaping the Gradient Vanishing: Periodic Alternatives of\\
Softmax in Attention Mechanism}
\vskip 0.2cm
\Large{\bf Appendix}
\end{center}
In this supplemental material, we provide more plots and solutions to support our argument. In addition, we also provide more 3D graphs and experiment results omitted from the paper.

\subsection{How the other elements in the input influence the gradient.}\label{a1}
\begin{figure}[H]
\centering
\includegraphics[width=\textwidth]{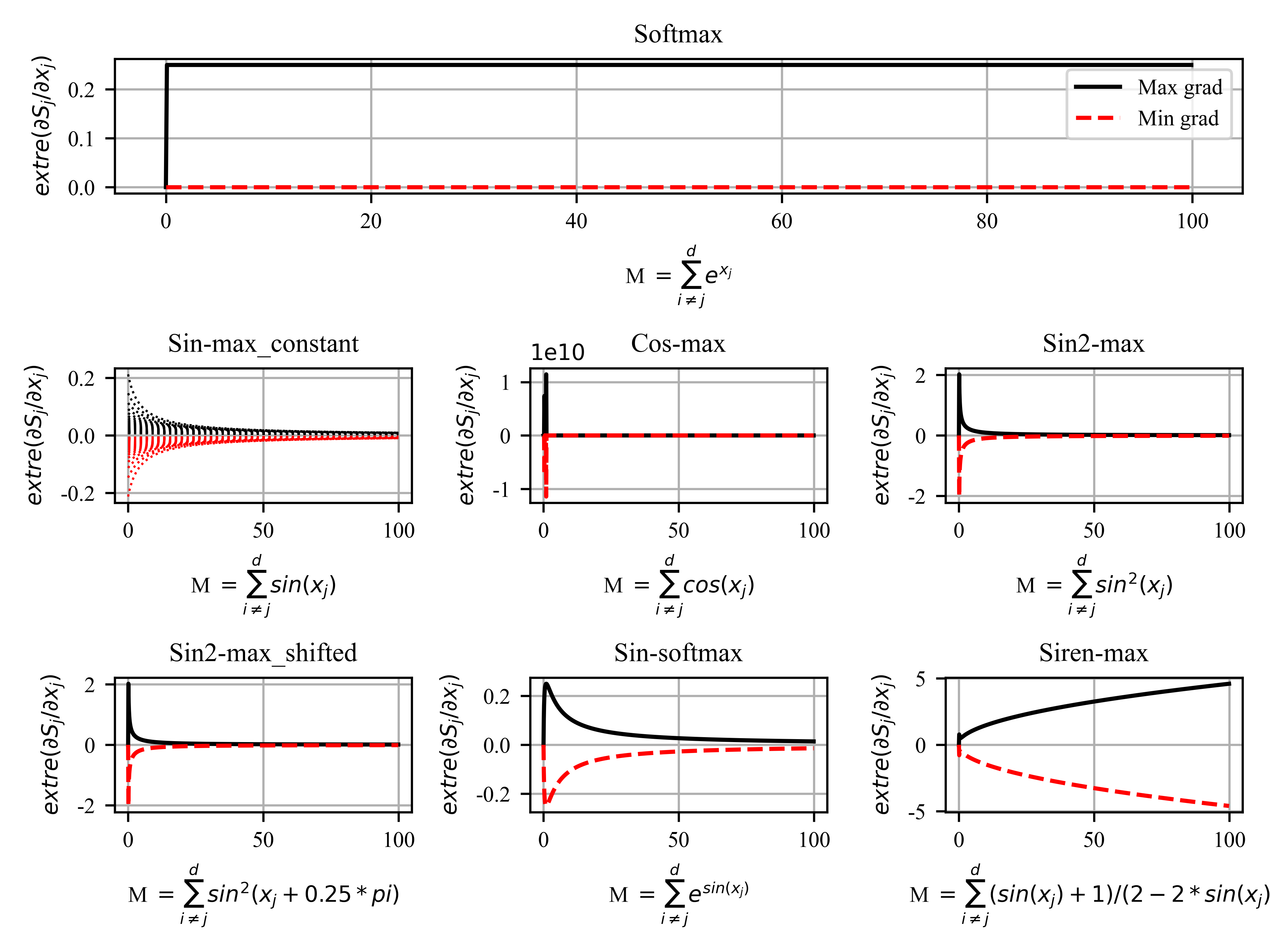}
\caption{The plot of the sum of $f(x)$ without $f\left(x_{j}\right)$ and the extreme gradient of $x_{j}$, where $d$ represents to the dimension of input. Since norm-Sin-max\_constant contains a constant term, there is an extra variable $d$ changing from $[2,256]$ with step\_size $=2$.}\label{fig8}
\end{figure}

Softmax and the periodic alternatives can be described as follows:
$$
S_{j}=\frac{f\left(x_{j}\right)}{f\left(x_{j}\right)+M^{\prime}}, M=\sum_{i \neq j}^{d} f\left(x_{i}\right)
$$
where $d$ represents to the dimension of input $x$.
$$
\frac{\partial S_{j}}{\partial x_{j}}=g(M) \cdot \frac{\partial f\left(x_{j}\right)}{\partial x_{j}}, g(M)=\frac{M}{\left(M+f\left(x_{j}\right)\right)^{2}}
$$
Since $f\left(x_{j}\right)$ is expected to be positive definite, we have $M \in(0,+\infty)$. The $g(M)$ can be considered as a band-pass filter with a maximum pass rate of $-\infty$, and the center frequency is $f\left(x_{j}\right)$. As the $M$ moves away from the center frequency, the pass rate gradually decays to $0$. Therefore, the closer $M$ is to $f\left(x_{j}\right)$, the more smoothly the gradient of $x_{j}$ can pass through the filter. In another words, the greater the difference between $f\left(x_{i}\right)$, the less smoothly the gradient flows. Accordingly, from the perspective of gradient, $f(x)$ should not be a function with sharp attenuation. Less attenuation is one of the reasons why periodic alternatives are superior to Softmax.

\subsection{pre-Normalization and gradient}\label{a2}
Normalization can be described as follows:
$$
\operatorname{norm}\left(x_{j}\right)=\frac{x_{j}-\operatorname{mean}\left(x_{j}\right)}{\sqrt{\operatorname{var}\left(x_{j}\right)}}
$$
$$
\frac{\partial \operatorname{norm}\left(x_{j}\right)}{\partial x_{j}}=\frac{d-1}{d \cdot \sqrt{\operatorname{var}\left(x_{j}\right)}}-\frac{x_{j}-\operatorname{mean}\left(x_{j}\right)}{2 \cdot \sqrt[3]{\operatorname{var}\left(x_{j}\right)}} \cdot \frac{\partial \operatorname{var}\left(x_{j}\right)}{\partial x_{j}}
$$
Denote the dimension of $x$ by $d$, then we have:
$$
\frac{\partial \operatorname{var}\left(x_{j}\right)}{\partial x_{j}}=\frac{d-1}{d^{2}} \cdot\left(x_{j}-\operatorname{mean}\left(x_{j}\right)\right)-\frac{1}{d^{2}} \sum_{i \neq j}^{d}\left(x_{i}-\operatorname{mean}\left(x_{j}\right)\right)
$$
$$
\operatorname{mean}\left(x_{j}\right)=x_{j}+\sum_{i \neq j}^{d} x_{i}
$$

\begin{figure}[H]
\centering
\includegraphics[width=0.9\textwidth]{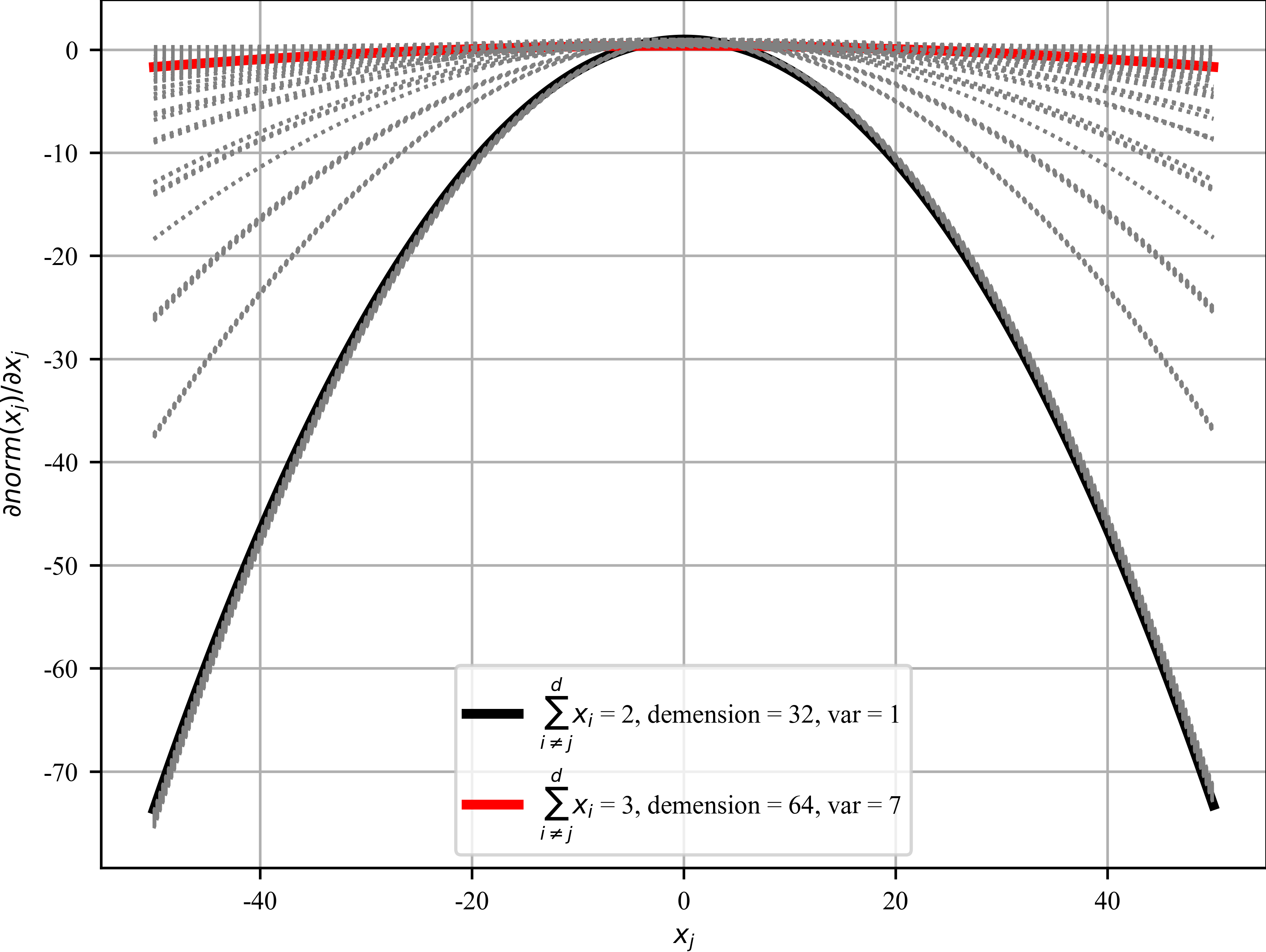}
\caption{The gradient plot of normalization. Note that there are three extra variables: dimension of input $d$, statistical variance of input $var$ and the sum of input without $x_{j}$, and we only annotate two sets of variables here.}\label{fig9}
\end{figure}

\begin{figure}[H]
\centering
\includegraphics[width=0.9\textwidth]{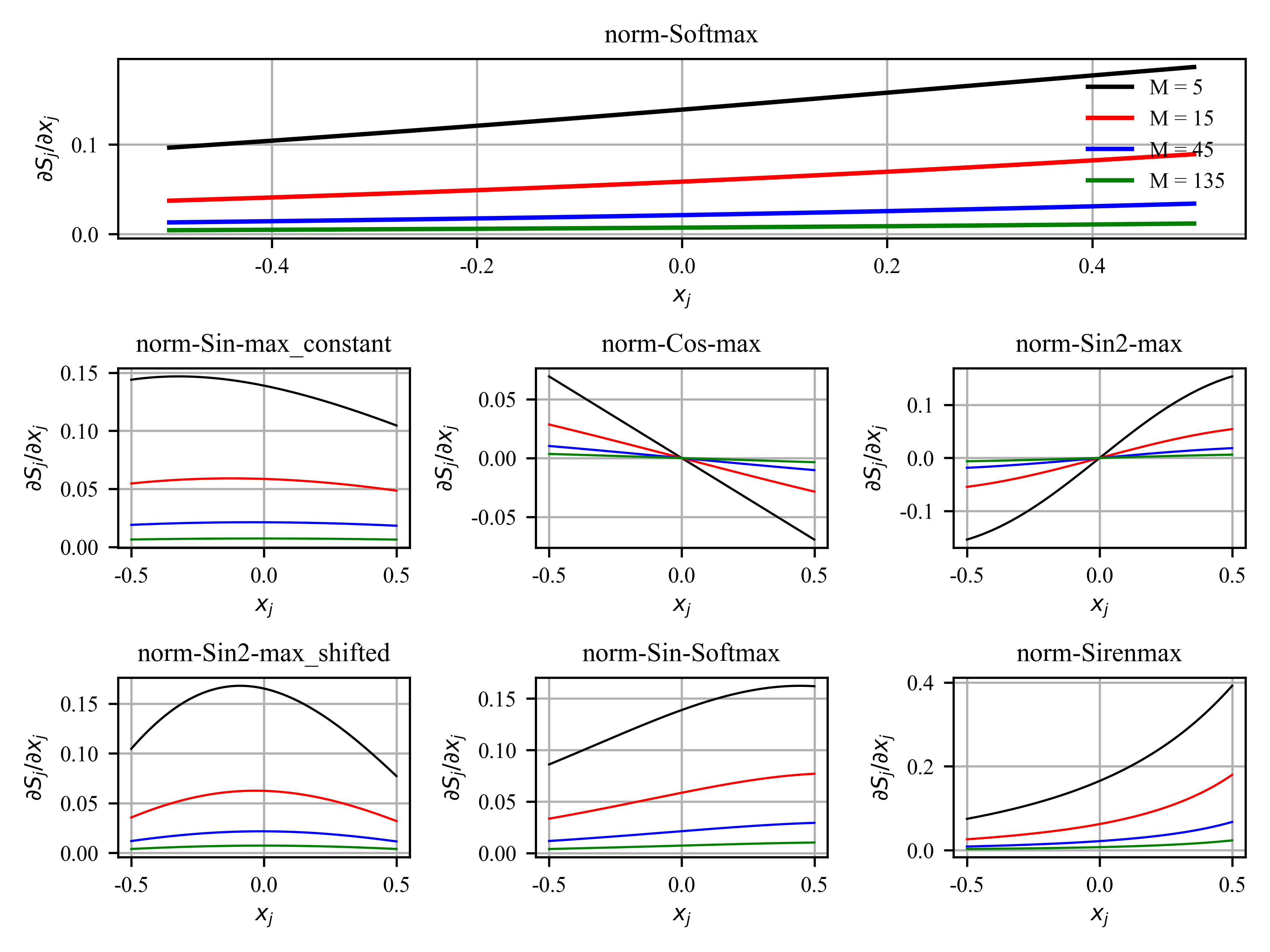}
\caption{The gradient plot of pre-normalized version of Softmax and periodic alternatives, where $M=\sum_{i \neq j}^{d} f\left(x_{i}\right)$, and $d$ represents to the dimension of input. For example, for norm-Sin-max\_constant, $M=\sum_{i \neq j}^{d} 1+\sin \left(x_{i}\right)$.}\label{fig10}
\end{figure}

Then we can plot the gradient of normalization as Figure \ref{fig9}. The gradient of normalization is related to the distribution of input, which might lead to extra gradient problems, such as gradient vanishing or explosion. Therefore, although pre-normalization can roughly gather the input into the best performed area of functions as shown in Figure \ref{fig10}, it is not always a good choice, but just a conditional solution depending on the distribution of input.

\begin{figure}[H]
\centering
\includegraphics[width=0.9\textwidth]{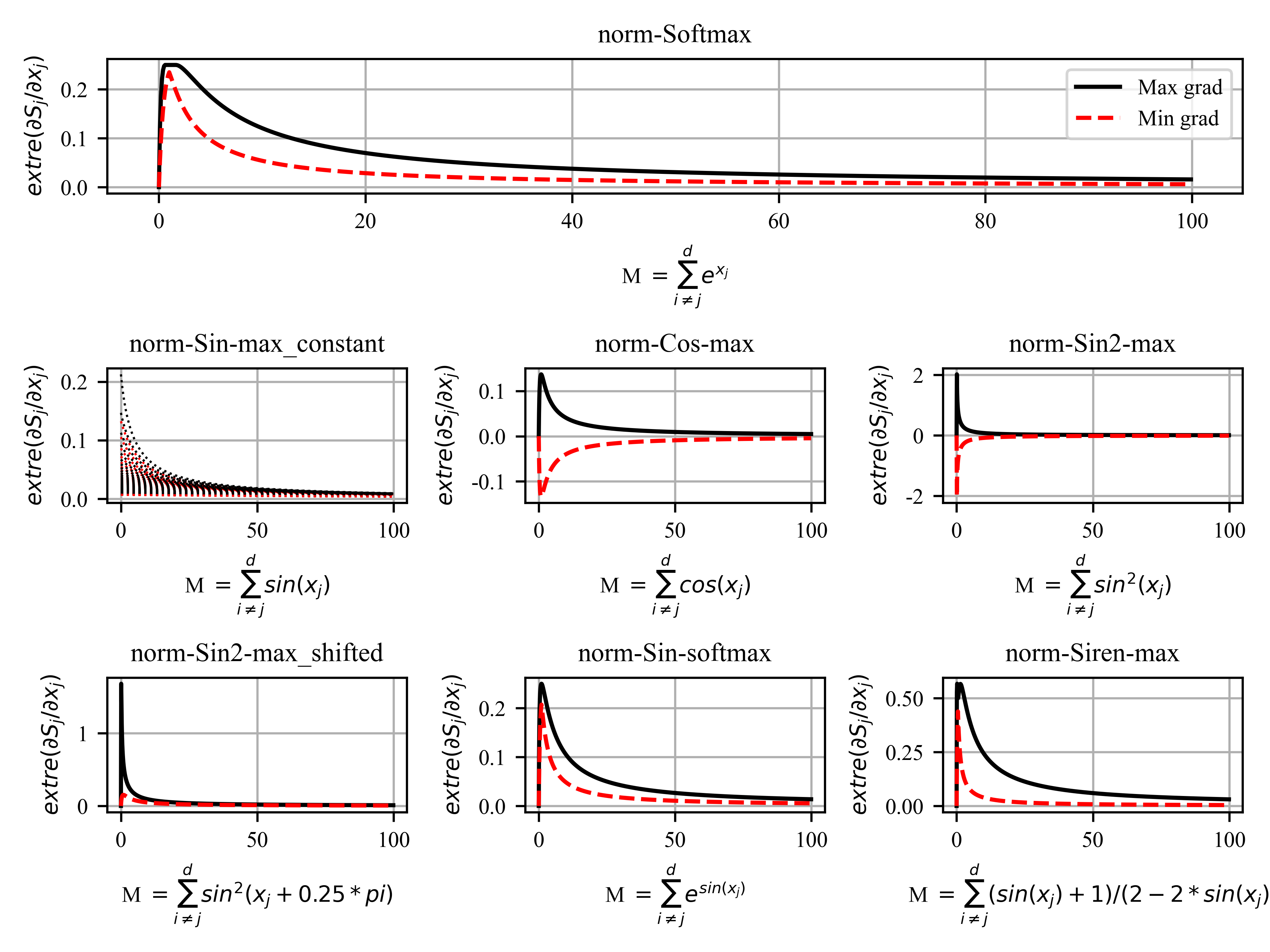}
\caption{The pre-normalized version of the plots on the sum of input without $x_{j}$ and the extreme gradient of $x_{j}$. Since norm-Sin-max\_constant contains a constant term, there is an extra variable $d$ changing from $[2,256]$ with step\_size $=2$, representing to the dimension of input.}\label{fig11}
\end{figure}

\subsection{3D graphs of gradients extracted from experiment}\label{a3}
\begin{figure}[H]
\centering
\includegraphics[width=0.75\textwidth]{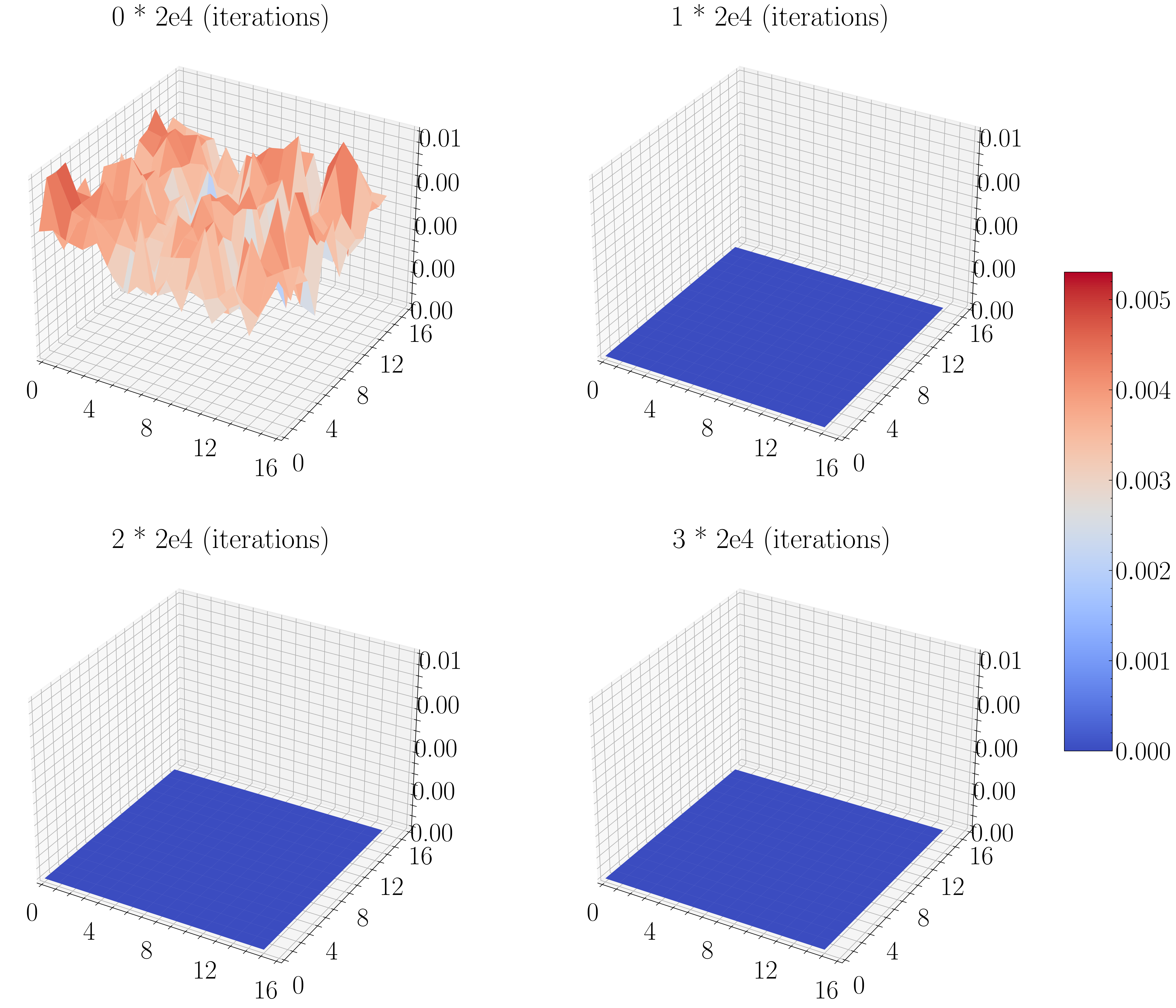}
\caption{3D gradient graph of Cos-max in experiment}\label{fig12}
\end{figure}

\begin{figure}[H]
\centering
\includegraphics[width=0.75\textwidth]{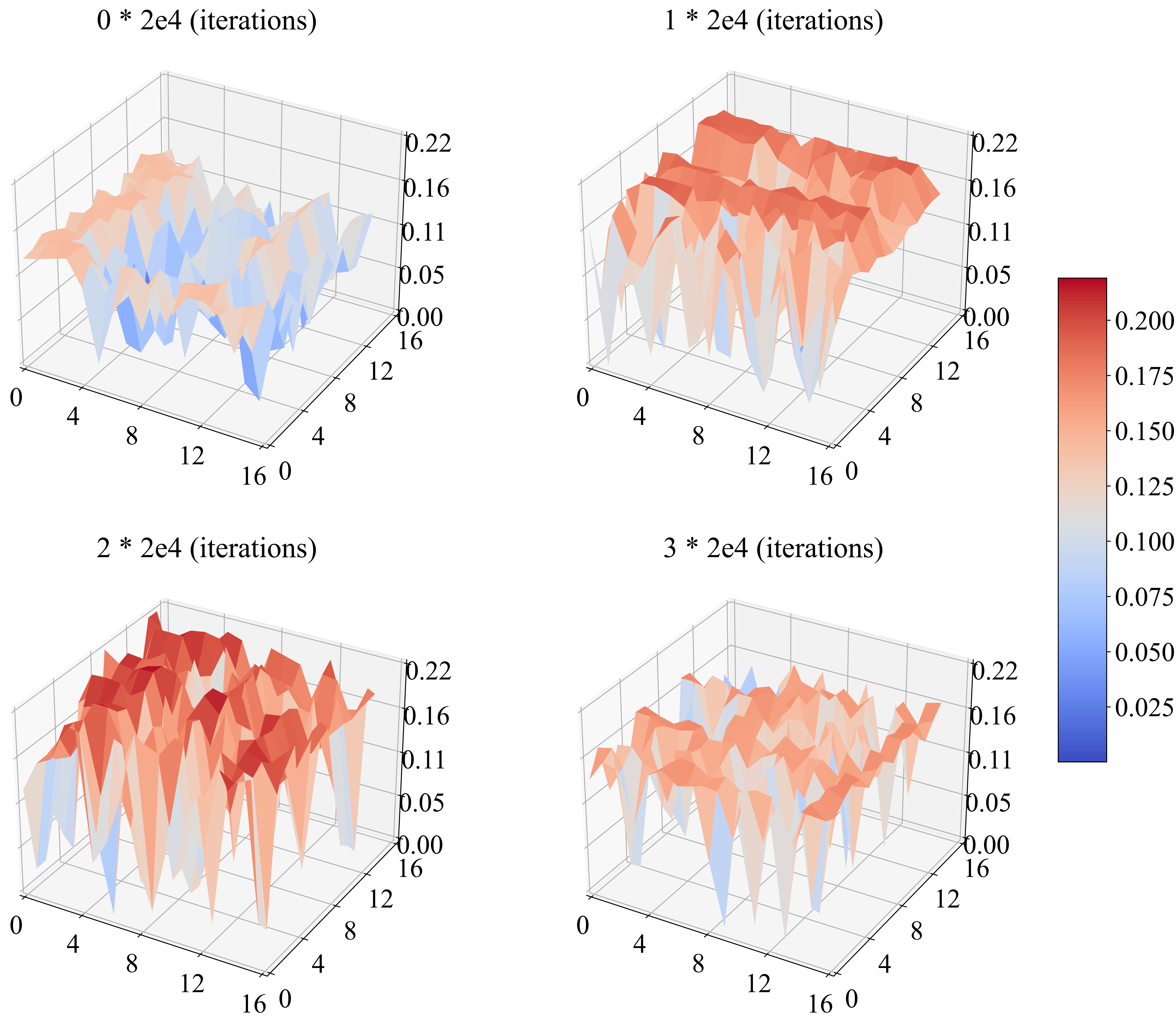}
\caption{3D gradient graph of Sin-Softmax in experiment}\label{fig13}
\end{figure}

\begin{figure}[H]
\centering
\includegraphics[width=0.85\textwidth]{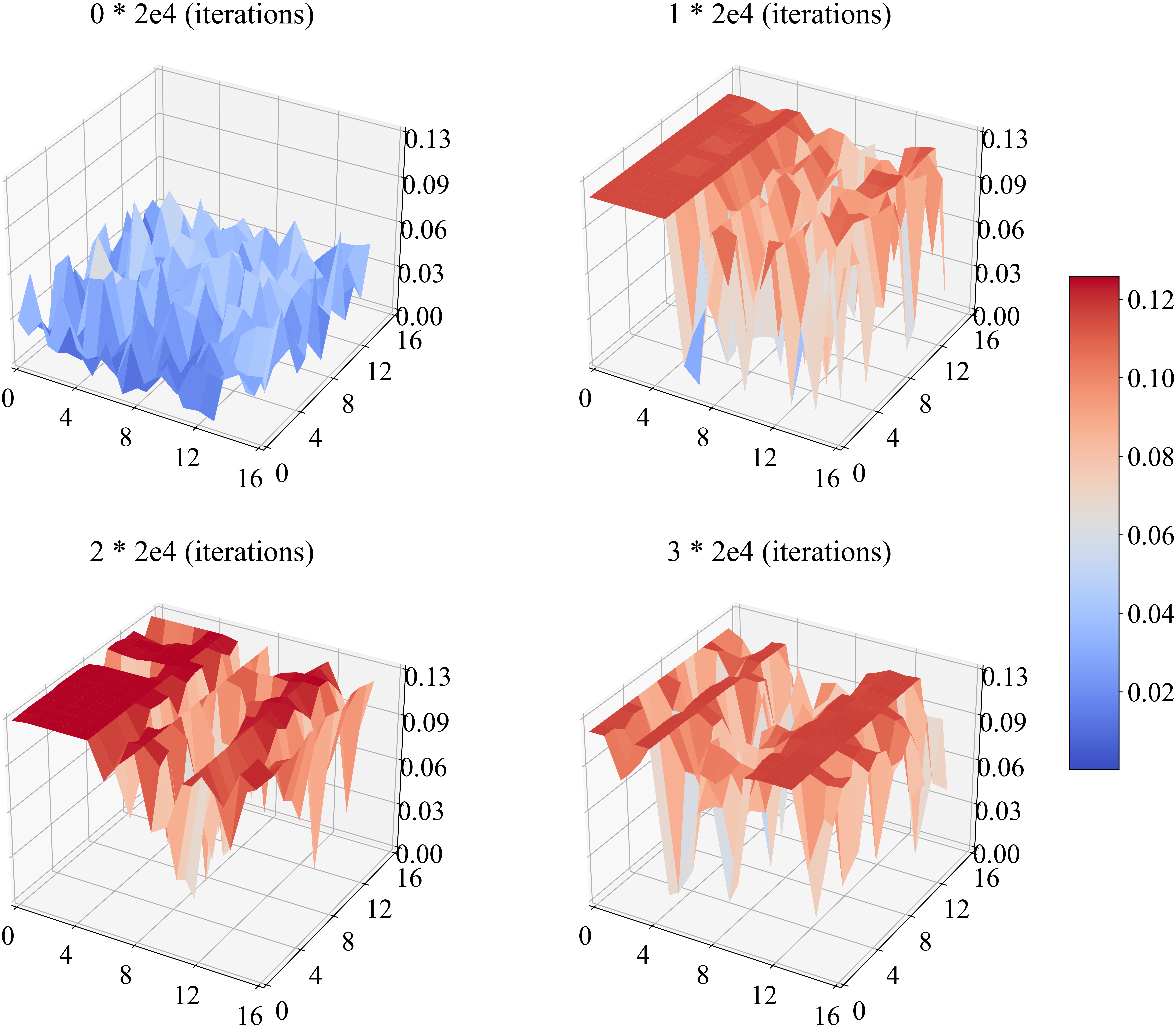}
\caption{3D gradient graph of Sin2-max-shifted in experiment}\label{fig14}
\end{figure}

As shown in Figure \ref{fig12}, just as we speculate, the gradient of Cos-max falls to zero due to the poor gradient performance in zero-region, and the gradient Sin2-max-shifted and Sin-Softmax is well performed at each stage of training as shown in Figure \ref{fig13} and Figure \ref{fig14}.

\begin{figure}[H]
\centering
\includegraphics[width=0.85\textwidth]{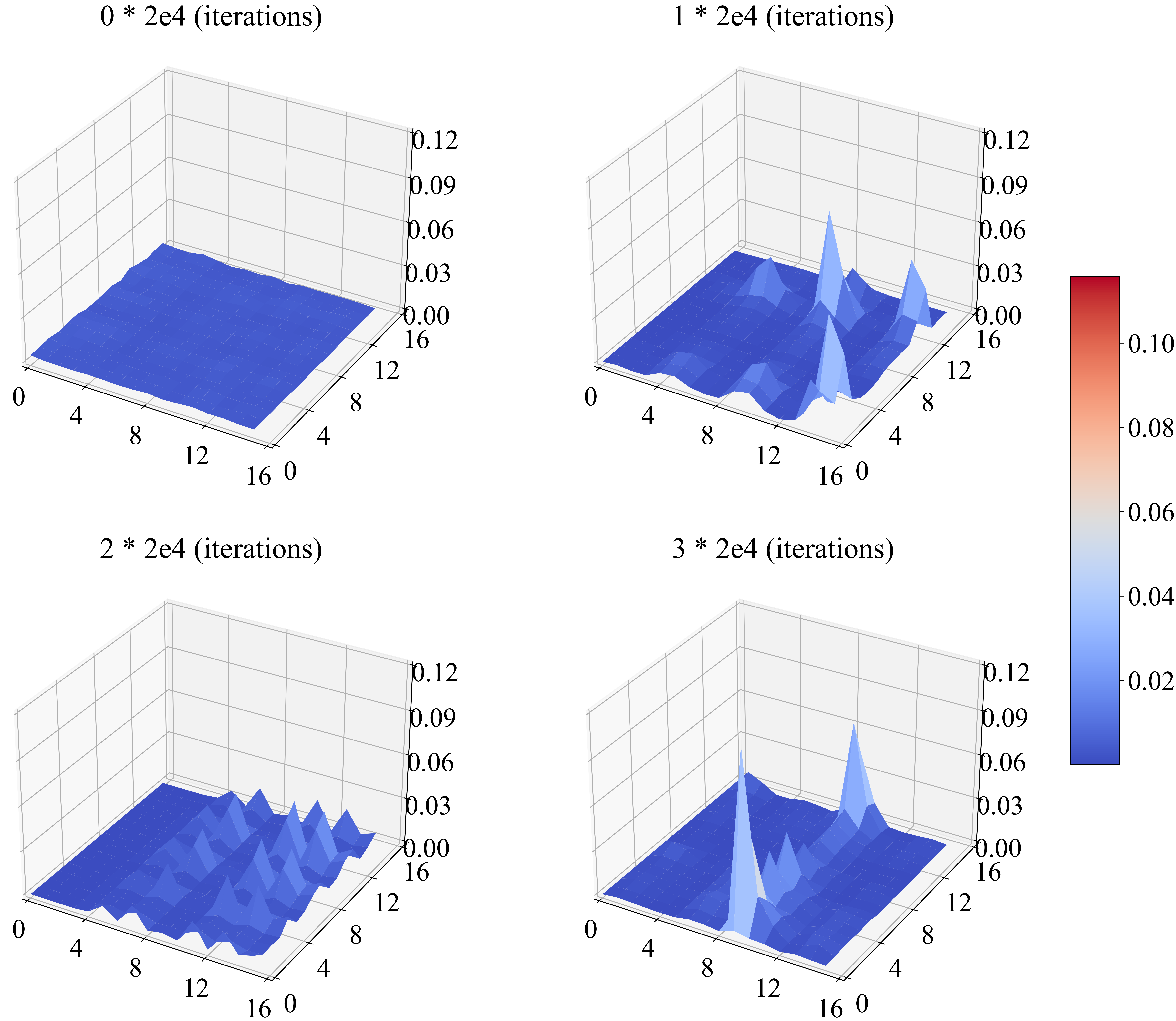}
\caption{3D gradient graph of Softmax in experiment}\label{fig15}
\end{figure}

\subsection{Experiment results}\label{a4}
\begin{figure}[H]
\centering
\includegraphics[width=\textwidth]{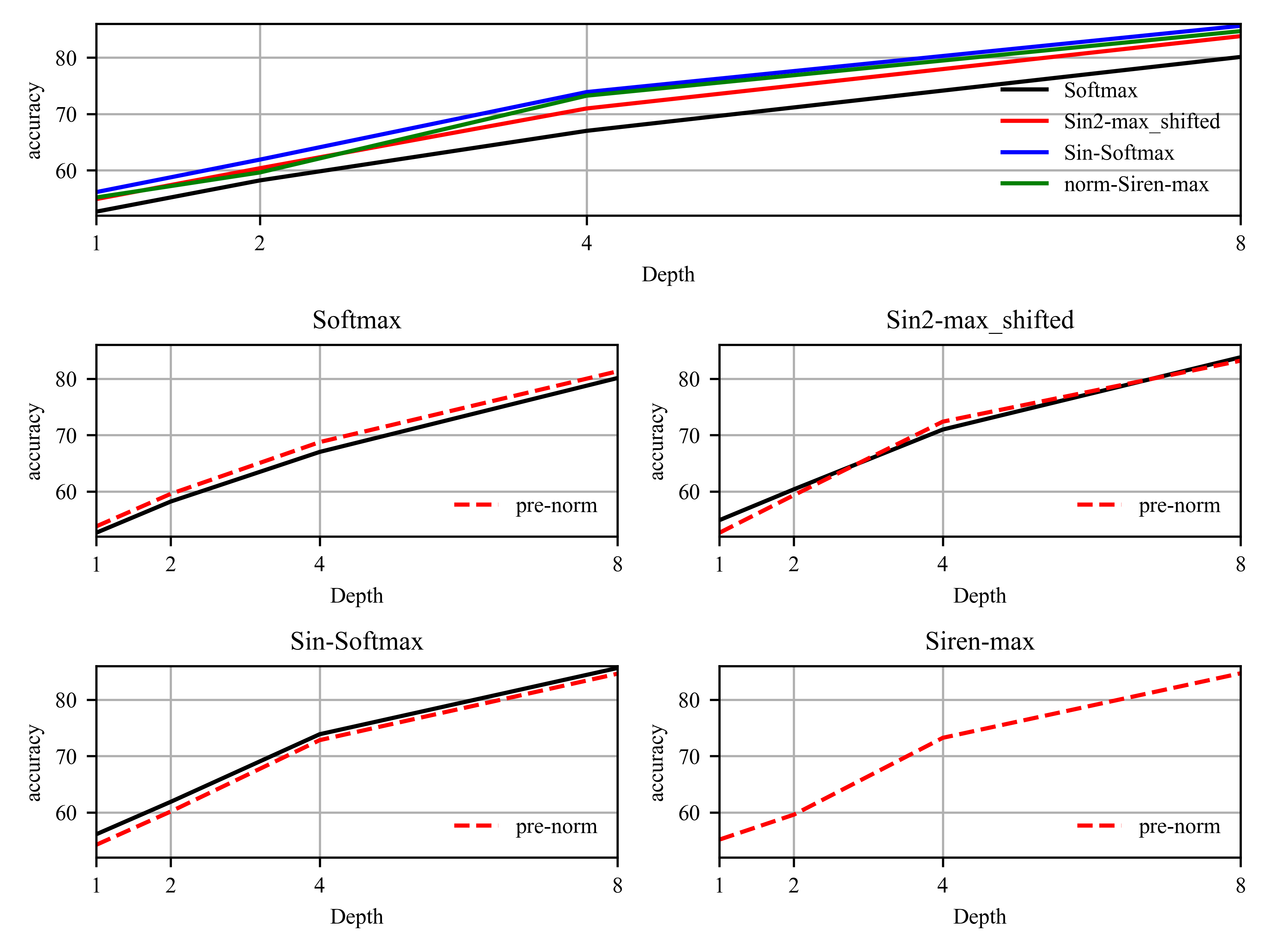}
\caption{Experiment results of Softmax, periodic alternatives and their pre-normalized version. The depth represents to the number of attention blocks stacked, and the rest structures of demo remain unchanged.}\label{fig16}
\end{figure}

\begin{table}[H]
\centering
\caption{Softmax and periodic alternatives are tested under increasing depth. The depth represents to the number of attention blocks stacked, and the rest structures of demo remain unchanged.}\label{tb3}
\begin{tabular}{ccccccccc}
\toprule
\multirow{2}{*}{\textbf{\begin{tabular}[c]{@{}c@{}}Structure = Demo\\ Dataset = Cifar-100\end{tabular}}} & \multicolumn{2}{c}{\textbf{Depth = 1}} & \multicolumn{2}{c}{\textbf{Depth = 2}} & \multicolumn{2}{c}{\textbf{Depth = 4}} & \multicolumn{2}{c}{\textbf{Depth = 8}} \\
                                                     \cline{2-9} &                    & norm              &                    & norm              &                    & norm              &                    & norm              \\\midrule
\textbf{Softmax}                                                                                         & 52.67              & 53.80             & 58.21              & 59.57             & 67.01              & 68.75             & 81.12              & 82.31             \\
\textbf{Sin-max-constant}                                                                                & \textbackslash{}   & \textbackslash{}  & \textbackslash{}   & \textbackslash{}  & \textbackslash{}   & \textbackslash{}  & \textbackslash{}   & \textbackslash{}  \\
\textbf{Cos-max}                                                                                         & 0.496              & 0.480             & 0.493              & \textbackslash{}  & \textbackslash{}   & \textbackslash{}  & \textbackslash{}   & \textbackslash{}  \\
\textbf{Sin2-max}                                                                                        & \textbackslash{}   & \textbackslash{}  & \textbackslash{}   & \textbackslash{}  & \textbackslash{}   & \textbackslash{}  & \textbackslash{}   & \textbackslash{}  \\
\textbf{Sin2-max-shifted}                                                                                & 53.90              & 53.16             & 60.37              & 59.84             & 71.48              & 72.38             & 84.81              & 83.20             \\
\textbf{Sin-Softmax}                                                                                     & \textbf{54.64}     & 54.27             & \textbf{60.40}     & 60.18             & 72.39              & 71.84             & \textbf{85.14}     & 84.63             \\
\textbf{Siren-max}                                                                                       & \textbackslash{}   & 55.19             & \textbackslash{}   & 59.62             & \textbackslash{}   & \textbf{73.25}    & \textbackslash{}   & 84.70             \\ \bottomrule
\multicolumn{9}{c}{\textbf{*  \textbackslash  ~means training breaks down in the early stages}}                                                                                                                                                                              
\end{tabular}
\end{table}
As shown in Table \ref{tb3} and Figure \ref{fig16}, due to the poor zero-region gradient performance, training under Sin2-max breaks down in the early stages, as same as Cos-max under a greater depth. The input of Sin-max-constant is submerged by the constant term, and the breaking down also occurs. In addition, Siren-max need normalization to make gradient avoid jump point, otherwise it perform poorly.

\subsection{Solutions}\label{a5}
{\bf Solution for the range of extremum gradient of Cos-max:}
$$
\frac{\partial S_{j}}{\partial x_{j}}=\frac{-M \cdot \sin \left(x_{j}\right)}{\left(M+\cos \left(x_{j}\right)\right)^{2}}
$$
$$
\frac{\partial^{2} S_{j}}{\partial x_{j}^{2}}=-M \cdot \cos \left(x_{j}\right) \cdot\left(M+\cos \left(x_{j}\right)\right)^{-2}+2 \cdot M \cdot \sin \left(x_{j}\right) \cdot\left(M+\cos \left(x_{j}\right)\right)^{-3}
$$
Let $\partial^{2} S_{j} / \partial x_{j}^{2}=0$, we have:
$$
\cos \left(x_{j}\right)-2 \cdot \sin \left(x_{j}\right) \cdot\left(M+\cos \left(x_{j}\right)\right)^{-1}=0
$$
$$
\sin \left(x_{j}\right) \cdot\left(M+\cos \left(x_{j}\right)\right)^{-2}=0.5 \cdot \cos \left(x_{j}\right) \cdot\left(M+\cos \left(x_{j}\right)\right)^{-1}
$$
$$
\therefore  \text { Extre }\left(\frac{\partial S_{j}}{\partial x_{j}}\right)=\frac{-M \cdot \cos \left(x_{j}\right)}{M+\cos \left(x_{j}\right)}=\frac{M^{2}}{2 M+2 \cos \left(x_{j}\right)}-\frac{M}{2}
$$
$$
\therefore \text { Extre }\left(\frac{\partial S_{j}}{\partial x_{j}}\right) \in\left[\frac{M^{2}}{2 M+2}-\frac{M}{2}, \frac{M^{2}}{2 M-2}-\frac{M}{2}\right]
$$
Since Cos-max is not positive definite, $M \in[-\infty,+\infty]$. So we have:
$$
\text { Extre }\left(\frac{\partial S_{j}}{\partial x_{j}}\right) \in(-\infty,+\infty)
$$

\end{document}